# Fountain – an intelligent contextual assistant combining knowledge representation and language models for manufacturing risk identification

Saurabh Kumar[$], Daniel Fuchs, Klaus Spindler

Artificial Intelligence Technologies, FORVIA Clean Mobility, Augsburg, Germany

[$]Primary author

Abstract

Deviations from the approved design or processes during mass production can lead to unforeseen risks. However, these changes are sometimes necessary due to changes in the product design characteristics or an adaptation in the manufacturing process. A major challenge is to identify these risks early in the workflow so that failures leading to warranty claims can be avoided. We developed Fountain as a contextual assistant integrated in the deviation management workflow that helps in identifying the risks based on the description of the existing design and process criteria and the proposed deviation. In the manufacturing context, it is important that the assistant provides recommendations that are explainable and consistent. We achieve this through a combination of the following two components 1) language models finetuned for domain specific semantic similarity and, 2) knowledge representation in the form of a property graph derived from the bill of materials, Failure Modes and Effect Analysis (FMEA) and prior failures reported by customers. Here, we present the nuances of selecting and adapting pretrained language models for an engineering domain, continuous model updates based on user interaction with the contextual assistant and creating the causal chain for explainable recommendations based on the knowledge representation. Additionally, we demonstrate that the model adaptation is feasible using moderate computational infrastructure already available to most engineering teams in manufacturing organizations and inference can be performed on standard CPU only instances for integration with existing applications making these methods easily deployable.

## 1. Introduction

In the automotive domain, it is common practice to specify and approve all the product design and production process characteristics before the start of mass manufacturing. Component suppliers acquire approval from the Original Equipment Manufacturer for these specifications and are expected to keep them unchanged during the mass production phase. However, there are scenarios when a product design characteristic (e.g. the material used, dimension or tolerances) or a production process (e.g. machines used, production sequence, or tooling) have to be changed after the mass production has started. Scenarios like change in the supplier of a subcomponent, unavailability of a material or need to add a manual production or inspection process are not uncommon making such changes unavoidable. Any such change can lead to unforeseen risks related to product quality and lead to warranty claims. These claims can sometimes be very expensive. Hence it is important to identify any risks originating from such changes very early during the workflow. Processes for tracking and documenting changes are common practice and are often implemented within quality control software applications. However, the processes rely heavily on the ability of the individuals responsible for requesting and approving these changes in the production plant to identify and mitigate the risks. This is exacerbated by the need to quickly find solutions and implement the changes and the variance in experience and knowhow across production plants spread all around the world. Hence, a solution that considers these human factors (Godwin & Ebiefung, 1999) to retain a simple process without compromising on the quality of risk assessment is required.

This is a scenario where an intelligent contextual assistant (Dhiman, Wächter, Fellmann, & Röcker, 2022), integrated within the quality control process can assist the change requestor in identifying the risks very early in the workflow. However, the development of such an assistance system faces two major challenges – 1) having the domain knowledge that enables the identification of risks related to design or process changes, and 2) ability to map the textual description of change to the correct risks in the presence of domain specific terms.

Here, we demonstrate how we integrated such a contextual assistant within the quality control process related to deviation management within our organization. As shown in Fig. 1, the deviation management application is a web application deployed in the cloud that enables documentation, approval and tracking of every manufacturing deviation.

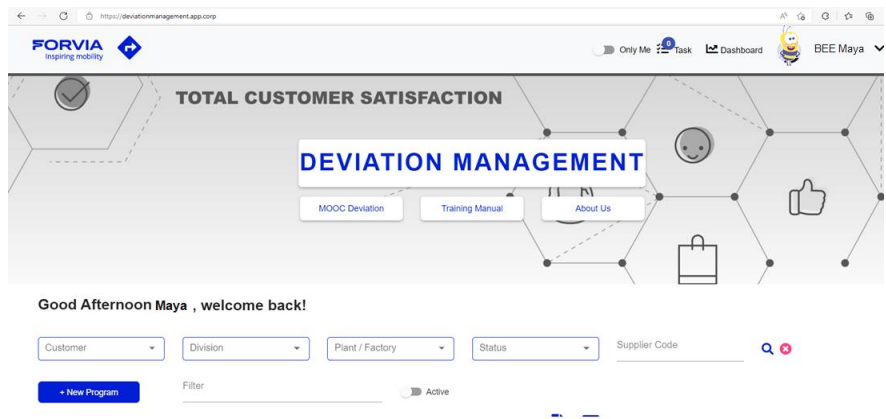

**Fig. 1** The deviation management application for the quality control process

As shown in Fig 2, the deviation requestor is required to provide the information related to the impacted part or assembly, a textual description of the current design or process and a textual description of the requested deviation.

**Fig. 2** The data to be provided for the deviation request

The goal of the contextual assistant is to help the deviation requestor in identification of the risks and checking whether the risk evaluation sufficiently covers all the risks. As shown in Fig. 3, the contextual assistance is required to use all this information and provide help with risk identification.

Fig. 3 The contextual assistance for risk identification provided by fountain

## 2. Methods

To achieve the goals of the contextual assistant, two main components are required.

1. A representation of the domain knowledge (Van Harmelen, Lifschitz, Porter, & eds, 2008) capturing causal relations between product and process characteristics and the failures that could emanate from changes in these characteristics
2. A language model (Bengio, Ducharme, & Vincent, 2000) finetuned for the domain and the task of determining and ranking the sematic similarity between the text related to deviations entered by the user and the product and process characteristics.

Apart from these, a simple preprocessing component has been utilized to account for domain specific use of certain terms and abbreviations that cannot be easily captured by the language model due to lack of model finetuning samples and lack of variance within the available samples. A simple example in the domain presented here is the term 'cat'. The term 'cat' in our domain is used to refer to a catalyst and not to an animal as would be identified by any pretrained language model. Such terms have been identified by collecting frequently used terms referring to part names and mapping them to the names as represented in the engineering Bill of Materials (BOM) for the product. For this example, a deviation text containing 'cat' would be assigned to a part representing a 'catalyst'.

### 2.1 The domain knowledge representation

It is difficult to separate the task of domain knowledge representation from its intended usage. For the purpose of developing Fountain, we avoided attempting to develop a general-purpose knowledge representation for the products covering the entire design, manufacturing, sourcing and assembly of all subcomponents and their variants. Instead, we decided to reuse the high-quality sources of information that are available for almost all products and subcomponents and that can be incrementally added. Our goal has been to design a method that can easily scale across different types of products across the automotive domain and possibly generalize it to other domains like aerospace and medical devices. Since our goal is to identify potential failures arising from changes in product design, we focus on creating our knowledge representation from standard information sources that can enable this.

Failure Modes and Effect Analysis (FMEA) is a very good source of information linking parts, processes, failures, causes, and detection and prevention mechanisms (Teng & Ho, 1996) and has been applied in many industries (Wu, Liu, & Nie, 2021). In the automotive industry, it is common practice to perform an FMEA for every new product or change to existing products or processes. In our context, we had access to two different types of FMEAs that are extensively used within our organization – Design FMEAs and Process FMEAs. We extracted and preprocessed 1193 Design FMEAs and 565 Process FMEAs. The preprocessing eliminated duplicates involving the same part/process – failure – cause chains.

The BOM provides us a hierarchical representation of all the subcomponents that together create a final product. It is common practice across industries to maintain the BOM in a Product Lifecycle Management (PLM) system. The BOM hierarchies extracted from the PLM system serve as a representation of all parts and their relationships in the final product variants and is an important component of our knowledge representation.

The D-FMEA provides the relationship between product design characteristics and failures that could emanate from the design criteria not being fulfilled. As shown in fig.4, each part in the product can be linked to one or more failure modes that can have one or more causes and effects respectively. As mentioned previously, the goal of our representation is to enable failure identification based on the textual description of the existing definition and the deviation and to provide explainability for all the recommendations using the causal chains in the representations.

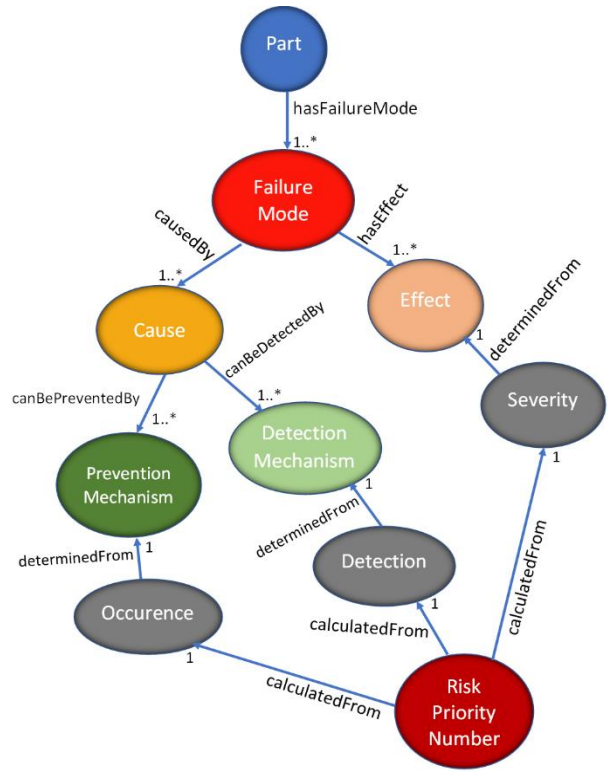

**Fig. 4** The D-FMEA concepts and their relationships

The representation has been instantiated as a labelled property graph (Robinson, Webber, & Eifrem, 2015) (Angles, 2018). A labelled property graph consists of nodes (having properties in the form of key-value pairs) and relationships. The relationships are labelled and directed and can have properties. There are several commercial and open-source frameworks that enable creation of property graphs. We have used the open-source framework *redisgraph* (Pieter, et al., 2019). It provides a simple containerized deployment and suits our cloud deployment scenario without relying on a proprietary managed service. *Cypher* (Francis, et al., 2018) has been used for graph querying and the query parameters can be dynamically generated depending on the contextual assistance scenario. Fig. 5 shows a sample subset of the property graph for one part (represented by the black circle) in the BOM.

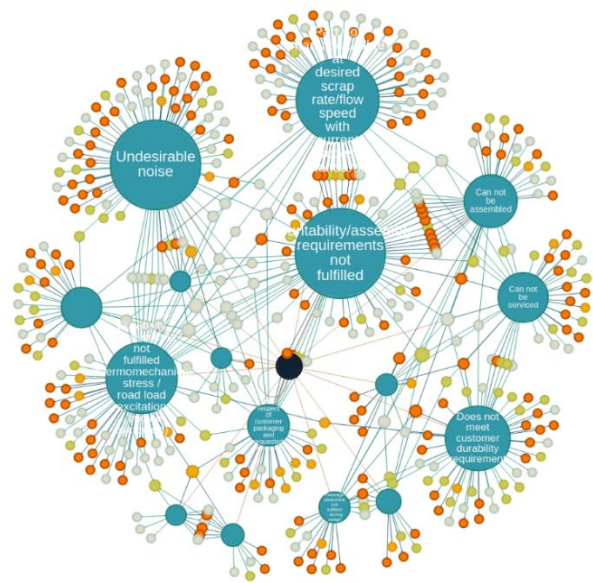

**Fig. 5** A subset of the graph showing the links between a single part (black dot), the failure modes (blue circles) and other concepts

In order to achieve the required contextual assistance, it is required to dynamically link the FMEA representation instances to the Deviation and Warranty Claim instances as shown in fig.6. Our two layered approach of knowledge representation with a layer of *Concepts* and the second layer of instantiation helps us achieve an overall representation that is popularly referred to as a knowledge graph (Barrasa, Hodler, & Webber, 2021) (Noy, et al., 2019) (Ji, Pan, Cambria, Marttinen, & Philip, 2021). Dynamically adding and linking instances from other domains related to product quality like Lessons Learnt and 8D Methodology can follow the same method and have been explained later in the section related to possible extensions.

To dynamically create the links between the concepts related to Deviation and Warranty Claims to the FMEA concepts as shown by the dotted lines in fig.6, it is required to determine the semantic similarity between the textual inputs related to deviation and warranty claims and the FMEA Causes, Effect and Detection Mechanisms. This is achieved using a domain adapted language model as explained in section 2.2.

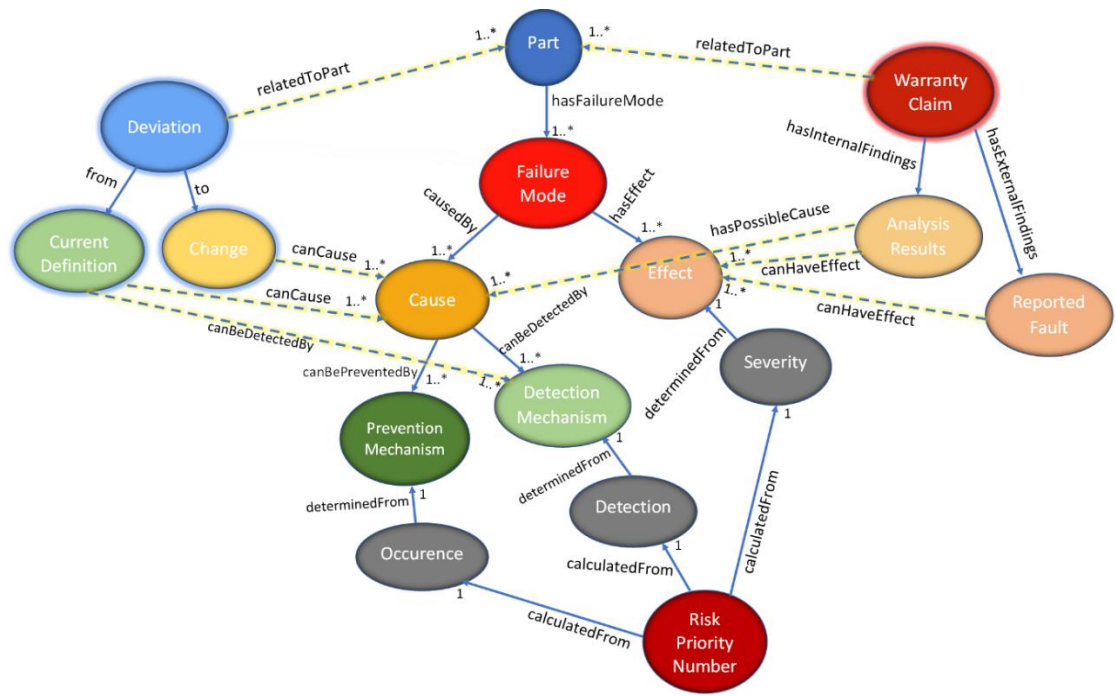

**Fig. 6** Dynamic addition and linking of nodes related to Deviation and Warranty Claims as shown by the dotted lines

2.2 The domain adapted language model

The advent of deep learning (LeCun, Bengio, & Hinton, 2015) in general, and the attention mechanism (Bahdanau, Cho, & Bengio, 2015) and the transformer architecture (Vaswani, et al., 2017) in particular, has led to tremendous advancements in natural language processing. BERT (Devlin, Chang, Lee, & Toutanova, 2018), RoBERTa (Liu, et al., 2019), XLNet (Yang, et al., 2019) and MPNet (Kaitao, Tan, Qin, Lu, & Liu, 2020) demonstrated that unsupervised pretraining followed by supervised fine tuning can enable large language models to achieve significant performance improvements on several benchmarks like STS-B (Daniel, Diab, Agirre, Lopez-Gazpio, & Specia, 2017), MNLI (Williams, Nangia, & Bowman, 2018) and MRPC (Dolan & Brockett, 2005). This has been followed by both creation of smaller models like DistilBERT (Sanh, Debut, Chaumond, & Wolf, 2019) and MiniLM (Wang, et al., 2020) achieving similar performance using distillation and much larger models with billions of parameters like GPT3 (Brown, et al., 2020), Gopher (Rae, et al., 2021) and BLOOM (Scao, et al., 2022) to achieve superior performance across a range of tasks.

The objective of dynamically linking deviation and warranty claim text to the FMEA text as mentioned in section 2.1 is an example of a semantic textual similarity task. The following two sentences that are very closely related in this context but sharing no common keywords demonstrate why semantics is so critical for this task:

i) *Presence of sharp areas lead to injury* and

ii) *Cone not safe to handle.*

These provide a good example to demonstrate why a method based on BM25 (Robertson & Zaragoza, 2009) that is used in several information retrieval solutions would provide inadequate results in such a scenario. Language models like RoBERTa or MPNet provide the feasibility to effectively handle such semantic similarity tasks. However, the biggest challenge in both scenarios is the need to simultaneously feed both sentences to the model in order to obtain the similarity score. This is computationally very expensive for an information retrieval task. There have been several attempts to generate sentence or paragraph level embeddings (Kiros, et al., 2015) (Conneau, Kiela, Schwenk, Barrault, & Bordes, 2017) (Cer, et al., 2018) (Reimers & Gurevych, 2019) with increasingly improved performance on the Semantic Textual Similarity (STS) tasks. This enables generation and storage of embeddings for a large corpus and calculating semantic similarity against a query for any information retrieval task. The approach from (Reimers & Gurevych, 2019) based on the modification of pretrained BERT and RoBERTa models using a Siamese network with a pooling layer on top significantly reduces the computational needs for training such a model. We evaluated the feasibility of using such a model with domain adaptation for the needs of our contextual assistance. Despite the current trend towards extremely large models requiring special GPU clusters, one of our goals has been to evaluate methods that require low computational costs during inference (execution on CPU only compute nodes in our Kubernetes clusters) and only moderate training costs (e.g. single GPU workstations). The low computational costs during inference enables us to create a highly responsive assistance feature as user is not expected to wait for the recommendations to show up. Another important motivation for our approach is to provide all engineering design and quality conformance teams the feasibility to train and test the models on standard compute infrastructure already available to them once we provide the base software modules, opening the possibility for many further applications without the need for GPU clusters to be allocated. We expect that this would make adapting the methods proposed by us significantly easier within manufacturing organizations of all sizes. Once the benefit of the assistance feature is proven, use of larger models and GPU clusters for better performance becomes a simpler task with just the need for scaling computational power with sufficient justification for the costs.

One of the important steps before performing any domain adaptation is to evaluate whether the pretrained models can already provide sufficient performance as required for the domain. A small set of common failure modes in our domain (and in general in a product manufacturing domain) can be used to demonstrate the feasibility of using these models in a semantic similarity task. To demonstrate this, we present here the results using two groups of sentences as shown in Table 1. The sentences have been chosen to be generic and easily understandable and can be replicated across multiple products. Some sentences have been intentionally added that do not describe a failure in order to understand the limitations of the models for our application where users often describe why a change would not lead to a failure. This check has been used as an indicator of model suitability and not as a validation method. For validations a larger labelled dataset was later used.

| Sentence Group 1 | | Sentence Group 2 | |
|---|---|---|---|
| S1_1 | Durability requirements not fulfilled | S2_1 | Assembly fails |
| S1_2 | Traceability requirements not fulfilled | S2_2 | Assembly fails before defined life |
| S1_3 | Temperature requirements not fulfilled | S2_3 | Welding joint cracked |
| S1_4 | Acoustic requirements not fulfilled | S2_4 | Radiant noise due to vibration |
| S1_5 | Mounting requirements not fulfilled | S2_5 | Thermal constraints on surrounding parts |
| S1_6 | Leakage requirements not fulfilled | S2_6 | Rust appears after a period of time |
| S1_7 | Connection requirements not fulfilled | S2_7 | Diameter of cone is too small and requires rework |
| S1_8 | Visual requirements not fulfilled | S2_8 | Thermal load is within limits |
| S1_9 | Weight requirements not fulfilled | S2_9 | Reduced flow noticed |
| S1_10 | Flow requirements not fulfilled | S2_10 | No impact on flow due to substitute part |

**Table 1** Sentence groups used to evaluate embedding quality on semantic similarity task

To quickly observe the performance of models on a domain specific semantic similarity task, the following checks as mentioned in Table 2 can be used

| Sentence pairs with expected **high** similarity | Sentence pairs with expected **low** similarity |
| --- | --- |
| {S1_1, S2_2}, {S1_1, S2_6} | {S1_3, S2_8} |
| {S1_3, S2_5} | {S1_1, S2_8} |
| {S1_4, S2_4} | {S1_10, S2_10} |
| {S1_10, S2_9} | {S1_8, S2_2}, {S1_8, S2_4}, {S1_8, S2_5}, {S1_8, S2_9} |

**Table 2** Quick qualitative check for the model performance on domain specific semantic similarity task

Using such a minimalistic check as the first step provides a quick qualitative measure of the usability of the models for a domain specific semantic similarity task as in the case of linking deviations to failure causes and the failure modes. If these checks do not provide the expected results, it is unlikely that a larger validation data set would demonstrate the suitability of the models.

Embeddings were generated and semantic similarity using cosine similarity was calculated for the sentence pairs. We used pretrained models that are based on RoBERTa, DistilBERT and MPNET that have been finetuned on close on 1 billion sentence pairs on multiple datasets as mentioned in the respective model cards at the Hugging Face model hub (mentioned in appendix 1) according to the same architecture as mentioned in (Reimers & Gurevych, 2019). The results are shown in fig 7.

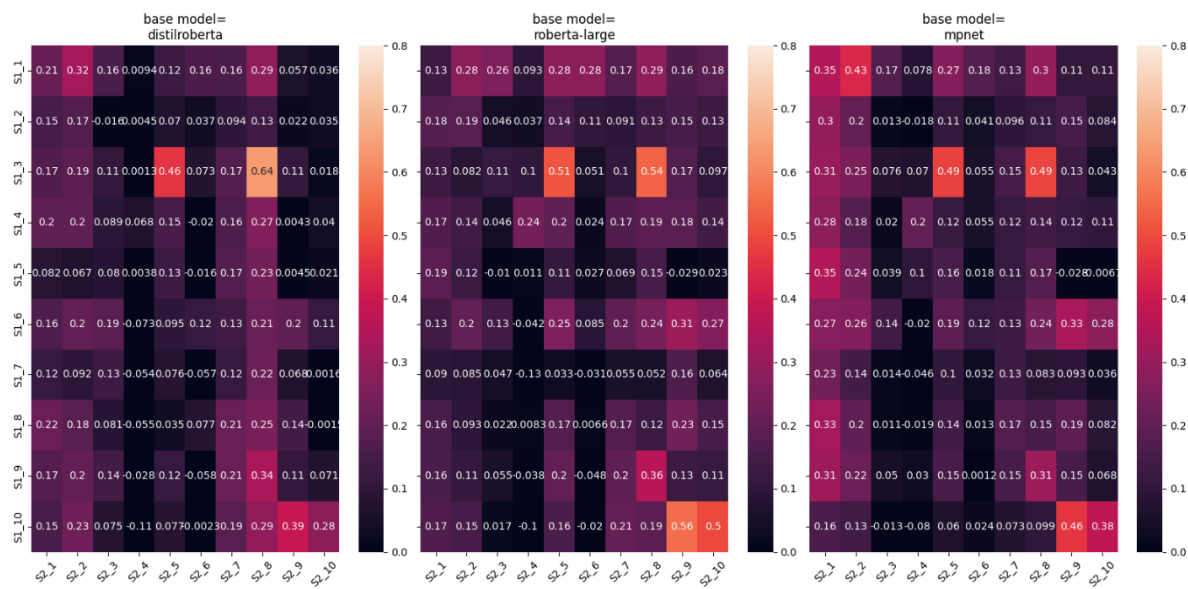

**Fig. 7** Cosine similarity for the sentence pairs in the sentence groups in Table 1

As can be seen, for all models, the highest cosine similarity is for the sentence pair {S1_3, S2_8} contrary to the expectation that the semantic similarity should be very low for this pair as mentioned in Table 2. This is also the case for multiple other sentence pairs like {S1_10, S2_10} where the actual semantic similarity is expected to be very low. This poses a major constraint for the usage of the models in our domain. Another observation is that the scores are too close to each other making it difficult for using thresholds to separate applicable and not applicable pairs. For example, separating {S1_4, S2_4} and {S1_4, S2_5} using the similarity scores would not be possible even though sentence pair {S1_4, S2_5} is not relevant. This makes the need for domain adaptation obvious.

As the first step in domain adaptation, we continued pretraining of the base model on domain data. The benefits of these, specially in a low resource setting, has been shown in (Gururangan, et al., 2020) (Zhu, et al., 2021). RoBERTa-large has been chosen as the base model and the continued pretraining was performed only for the masked language modelling task. This was followed by fine-tuning on the concatenated SNLI (Bowman , Angeli, Potts, & Manning, 2015) and MultiNLI (Williams, Nangia, & Bowman, 2018) datasets and evaluation on the STS tasks as proposed by (Reimers & Gurevych, 2019).

This was followed by finetuning using domain specific labelled dataset. To overcome the limitation of the models previously evaluated in handling negation as demonstrated above in the examples with sentence pairs {S1_3, S2_8} and {S1_10, S2_10}, a small set of sentences with negation were added to the finetuning dataset. The results for semantic similarity score calculations using the model at different stages for the sentence groups in Table 1 can be seen in fig 8. Fig 8(a) represents the performance of the model based on RoBERTa-large as in fig 7. Fig 8(b) shows the results with domain pretraining followed by only finetuning using SNLI and MultiNLI datasets. Fig 8(c) shows the results when additional finetuning using domain specific data is performed.

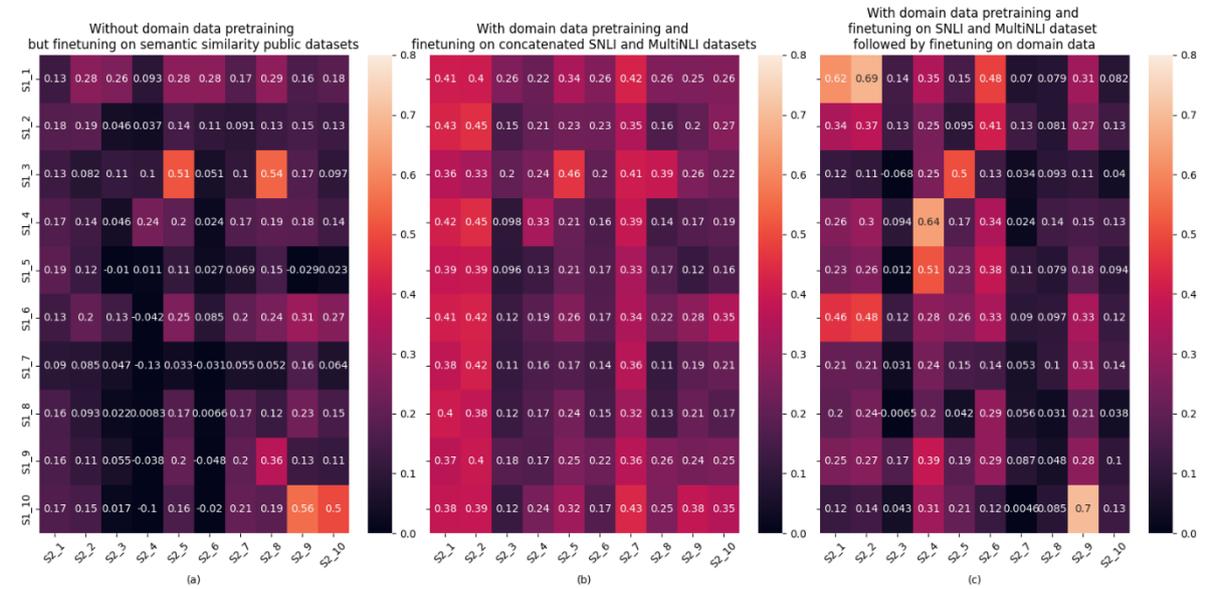

**Fig 8** (a) Model based on RoBERTa-large available on Hugging Face model hub trained using the methods in (Reimers & Gurevych, 2019) on approximately 1 billion sentence pairs, (b) Model based on RoBERTa-large with continued pretraining using domain data followed by finetuning on SNLI and MultiNLI, (c) The model mentioned in (b) with additional finetuning on domain data including negation

The results in fig 8 demonstrate the advantages of continued pretraining of the base model followed by finetuning on the domain dataset after the model has been finetuned on the publicly available dataset. The high difference in the cosine similarity scores between similar and dissimilar concepts make it feasible to use thresholds for separation of relevant sentence pairs. The model is also effectively able to deal with negation. To emphasize the need for handling negation and demonstrate the ability of the final model to do this, another small set of sentence groups is presented in Table 2. These sentence pairs provide an additional quick qualitative measure of the usability of the models in the presence of negation.

| Sentence Group 1 | | Sentence Group 2 | |
|---|---|---|---|
| S3_1 | Durability requirements not fulfilled | S4_1 | Assembly fails before defined life |
| S3_2 | Durability requirements satisfied | S4_2 | Welding joint cracked |
| S3_3 | Acoustic requirements not fulfilled | S4_3 | Radiant noise due to vibration |
| S3_4 | Acoustic requirements are met | S4_4 | Thermal constraints on surrounding parts |
| S3_5 | Leakage requirements not fulfilled | S4_5 | Sufficient sealing available |
| S3_6 | No leakage problems observed | S4_6 | Rust appears after a period of time |
| S3_7 | Flow requirements not fulfilled | S4_7 | Reduced flow noticed |
| S3_8 | Flow is as expected in design | S4_8 | No impact on flow due to substitute part |

**Table 3** Sentence groups to highlight the significance of the ability to handle negation

Fig 9 shows the efficacy of the model in dealing with similarity and negation together. The low cosine similarity score for pairs {S3_2, S4_1}, {S3_4, S4_3} and {S3_8, S4_7} demonstrate the ability to handle negation. The important result to notice is that the high semantic similarities are still retained in the presence of negation as the cosine similarity scores are high for the pairs {S3_6, S4_5}, {S3_8, S4_5} and {S3_8, S4_8}.

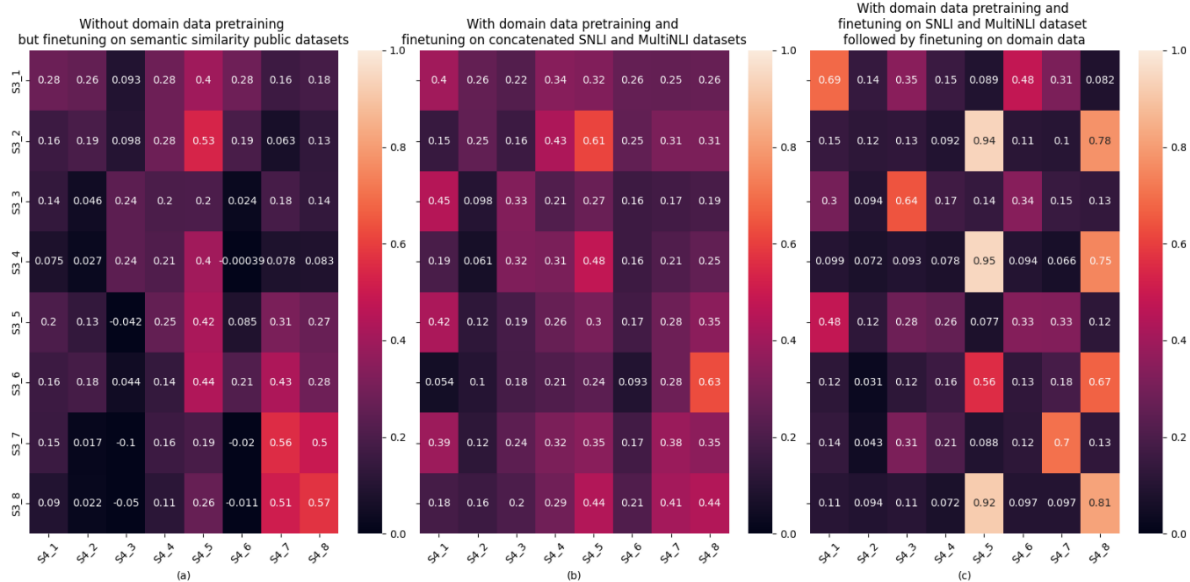

**Fig 9** (a) Model based on RoBERTa-large available on Hugging Face model hub trained using the methods in (Reimers & Gurevych, 2019) on approximately 1 billion sentence pairs, (b) Model based on RoBERTa-large with continued pretraining using domain data followed by finetuning on SNLI and MultiNLI, (c) The model mentioned in (b) with additional finetuning on domain data including negation

2.3 Assistance by combining the domain representation and the language model

As mentioned in section 1, the goal of the assistance feature is to enable the users in identifying the possible failures when they initiate the workflow for a manufacturing deviation. The language model is used to identify the failure causes that are available in the domain representation based on the semantic textual similarity to the user's deviation text. This is then used to create links between the deviation and the Failure Modes for the relevant parts as shown in fig 6. Additionally, this is used to identify the past warranty claims that could have a relationship to the failures and causes identified for the particular deviation. The user has the possibility to select the failures and the warranty claims that she/he considers relevant for the particular deviation as shown in fig 10. The user feedback is used as further data for model finetuning and performance evaluation in the live system.

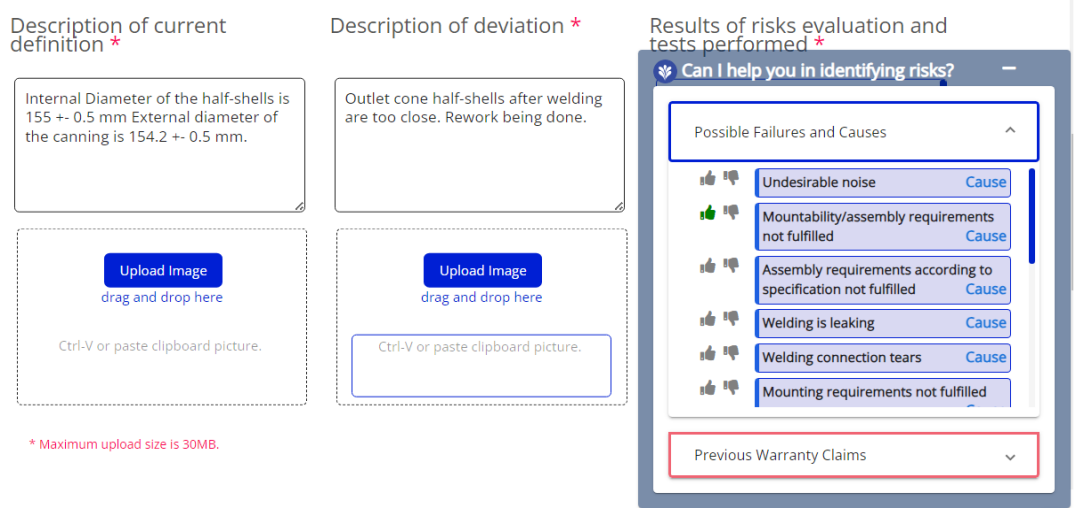

**Fig 10** The assistance feature provided by Fountain integrated within the deviation management application.

The users have the possibility to look at the details of the causes which could possibly lead to the respective failures as a consequence of this deviation as shown in fig 11. This provides explainability for the recommendations and helps the user in analyzing whether the failure can occur or not.

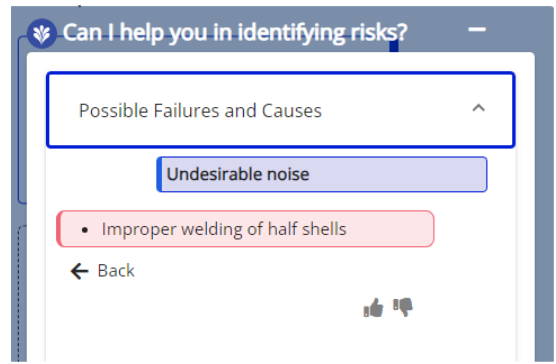

**Fig 11** Causes for the shown failures can be seen to help with the analysis

The like (thumbs up) and dislike (thumbs down) features were developed to perform anonymized user tests for the deployed application. Based on requests by the users during trials, this was additionally extended by adding the considered failure risks to the risk evaluation text with the users having the option to mention why this risk has already been considered and is not relevant and hence deviation approval can be obtained as shown in fig 12.

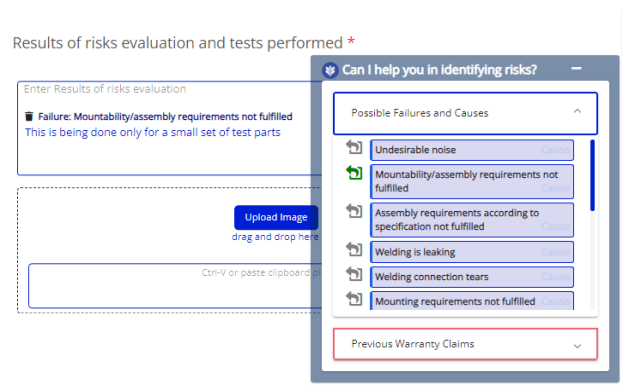

**Fig 12** The Failure text added to the risk evaluation when user selects it from the recommendation. The user can add the justification text below it for further approval process.

## 3. User study results

One of the major challenges is to validate the quality of recommendations for such an assistance system when it is integrated into another application. Unlike an ecommerce application, the number of users is small and a random selection of a small user subset who can preview and validate such an application is not feasible. Not every user has the same level of domain knowledge and expertise to validate the quality of recommendations. Another constraint in the low number of deviations that are created. The assistance system has to undergo detailed validation using a small user base and a small set of sample deviations before it can be rolled out to all the users across multiple regions and production plants. Hence, we performed tests in two stages – 1) with a set of three voluntary expert users who provided detailed results regarding the recommendations for multiple deviations and 2) anonymized tests where the results were calculated just based on like and dislike buttons with another set of expert users.

It is important to understand that not every recommendation for each deviation must be always fully applicable. This is very difficult, if at all possible, to achieve for a complex domain. Hence, we asked the expert users to rate each recommendation item as applicable or not applicable and obtained the following summary statistics – 1) number of deviations evaluated, 2) number of recommendations where all recommendation items were considered applicable, 3) number of recommendations where some recommendation items were considered applicable and some not applicable, and 4) number of recommendations where no recommendation items were considered useful. The result is shown in Table 4.

| User | Deviations evaluated | All recommendations useful | Both useful and non-useful recommendations | No useful recommendations |
|---|---|---|---|---|
| 1 | 59 | 29 | 22 | 8 |
| 2 | 11 | 4 | 3 | 4 |
| 3 | 7 | 0 | 4 | 3 |

**Table 4** Result of recommendation evaluation by selected expert users

The results in table 4 demonstrate that for a majority of deviations created, at least some of the recommendations from the assistance system are found useful and applicable by the users. Additionally, there are deviations where all recommendations from the assistance system are found useful by the end users. This is very positive for the high specificity of the system which is important for such an application and demonstrates the effectiveness of language model adaptation.

Additionally, the next stage of anonymized tests demonstrated that 34 recommendations were considered useful and 20 not useful. This is slightly inferior to the above results, but further evaluation of the data showed that some recommendations marked not useful contradict each other. Since the data has been anonymized in a way that they cannot be traced back to users, it is not feasible to evaluate whether the same user provided this feedback while experimenting with the system. We have considered this as a limitation of this anonymization approach of feedback collection and would attempt to implement a different approach without compromising user privacy in the next user study and in the productive live system.

## 4. Extensions and future work

The results demonstrate the effectiveness of using an intelligent assistant for such an application and acceptance of end users for such a system. In near future we intend to extend this method to problem solving methods like like Eight Disciplines of Problem Solving (8D) and continuous improvement methods like Kaizen and Lessons Learnt and Best Practices. This would involve adapting the knowledge representation for additionally linking data from these applications and reuse of the domain adapted language model. Additionally, we want to enable multiple teams to train and evaluate their own domain specific language models for other engineering and manufacturing applications and benefit from our approach.

## 5. Acknowledgements

We had excellent support from our colleagues from the group Total Customer Satisfaction and Quality at FORVIA in evaluating the effectiveness of the methods and results. This close cooperation enabled us to iteratively improve the user experience and the quality of recommendations. We would like to thank our colleague Vincent Besancon for preparing the deployment infrastructure in Azure Kubernetes Service and the model training infrastructure in Azure ML. Our colleague Ashwini Thanigavelu deserves credit for creating the scripts to extract and clean the FMEA data and our student intern Stephan Wolfgang Heinicke for extracting and translating the data related to deviations and warranty claims. We would also like to thank our colleague Brahmeshwar Reddy who made the required UI adaptations in the deviation management application to enable the assistance feature. Our colleague Daniel Saeuberlich deserves the credit for the name Fountain to depict the source of knowledge to quench the thirst of multiple application domains. He also helped in evaluating the correctness of the extracted FMEA data.

## Statements and declarations

This work has been carried out in the department Artificial Intelligence Technologies at FORVIA Clean Mobility, Augsburg, Germany. The FMEA, Deviation and Warranty Claim data are proprietary data.

## Author contributions

Saurabh Kumar created the concept of combining the knowledge representation in the form of property graph capturing the cause-effect relations with a language model to provide the assistance feature. He did the model training and finetuning required for domain adaptation of the language model. He also wrote the code for the microservices that constitute the application (Fountain) deployed in Azure Kubernetes Service and the MLOps pipelines and visualizations for training and evaluating performance of the different models.

Daniel Fuchs provided the domain knowledge and performed the labelling required for supervised finetuning. He created the list of domain specific synonyms and provided descriptions of domain specific terms. He also performed evaluations of the recommendation quality before user tests.

Klaus Spindler provided the ideas and suggestions for effectiveness of the cause-effect chains and the usefulness in the manufacturing and product design context. He performed multiple reviews for the application.

## 6. References

Angles, R. (2018). The Property Graph Database Model. *Proceedings of the 12th Alberto Mendelzon International Workshop on Foundations of Data Management.*

Bahdanau, D., Cho, K., & Bengio, Y. (2015). Neural Machine Translation by Jointly Learning to Align and Translate. *3rd International Conference on Learning Representations.* Retrieved from http://arxiv.org/abs/1409.0473

Barrasa, J., Hodler, A., & Webber, J. (2021). *Knowledge Graphs.* O'Reilly Media, Incorporated.

Bengio, Y., Ducharme, R., & Vincent, P. (2000). A neural probabilistic language model. *Advances in neural information processing systems.*

Bowman, S., Angeli, G., Potts, C., & Manning, C. (2015). A large annotated corpus for learning natural language inference. *Proceedings of the 2015 Conference on Empirical Methods in Natural Language Processing* (pp. 632-642). Association for Computational Linguistics.

Brown, T., Mann, B., Ryder, N., Subbiah, M., Kaplan, J., Dhariwal, P., & Neelakantan, A. (2020). Language models are few-shot learners. *Advances in neural information processing systems*, *33*, pp. 1877-1901.

Cer, D., Yang, Y., Kong, S.-y., Hua, N., Limtiaco, N., St. John, R., . . . Kurzweil, R. (2018). Universal Sentence Encoder for English. *Proceedings of the 2018 Conference on Empirical Methods in Natural Language Processing: System Demonstrations* (pp. 169-174). Association for Computational Linguistics.

Conneau, A., Kiela, D., Schwenk, H., Barrault, L., & Bordes, A. (2017). Supervised Learning of Universal Sentence Representations from Natural Language Inference Data. *Proceedings of the 2017 Conference on Empirical Methods in Natural Language Processing.*

Daniel, C., Diab, M., Agirre, E., Lopez-Gazpio, I., & Specia, L. (2017). Semeval-2017 task 1: Semantic textual similarity-multilingual and cross-lingual focused evaluation. *11th International Workshop on Semantic Evaluations (SemEval-2017)* (pp. 1–14). Vancouver, Canada: Association for Computational Linguistics.

Devlin, J., Chang, M.-W., Lee, K., & Toutanova, K. (2018). *Bert: Pre-training of deep bidirectional transformers for language understanding.* Retrieved from arXiv preprint: https://arxiv.org/abs/1810.04805

Dhiman, H., Wächter, C., Fellmann, M., & Röcker, C. (2022). Intelligent assistants: conceptual dimensions, contextual model, and design trends. *Business & Information Systems Engineering, 64*(5), 645-665.

Dolan, W., & Brockett, C. (2005). Automatically Constructing a Corpus of Sentential Paraphrases. *Proceedings of the Third International Workshop on Paraphrasing (IWP2005).* Retrieved from https://aclanthology.org/I05-5002

Francis, N., Green, A., Guagliardo, P., Libkin, L., Lindaaker, T., Marsault, V., . . . Taylor, A. (2018). Cypher: An evolving query language for property graphs. *Proceedings of the 2018 international conference on management of data*, (pp. 1433-1445).

Godwin, U. G., & Ebiefung, A. A. (1999). Human factors affecting the success of advanced manufacturing systems. *Computers & Industrial Engineering, 37*((1-2)), 297-300.

Gururangan, S., Marasović, A., Swayamdipta, S., Lo, K., Beltagy, I., Downey, D., & Smith, N. (2020). Don't Stop Pretraining: Adapt Language Models to Domains and Tasks. *Proceedings of the 58th Annual Meeting of the Association for Computational Linguistics* (pp. 8342-8360). Association for Computational Linguistics.

Ji, S., Pan, S., Cambria, E., Marttinen, P., & Philip, S. (2021). A survey on knowledge graphs: Representation, acquisition, and applications. *IEEE transactions on neural networks and learning systems, 33*(2), 494-514.

Kaitao, S., Tan, X., Qin, T., Lu, J., & Liu, T.-Y. (2020). Mpnet: Masked and permuted pre-training for language understanding. *Advances in Neural Information Processing Systems 33*, (pp. 16857-16867).

Kiros, R., Zhu, Y., Salakhutdinov, R., Zemel, R., Urtasun, R., Torralba, A., & Fidler, S. (2015). Skip-thought vectors. *Advances in neural information processing systems 28.*

LeCun, Y., Bengio, Y., & Hinton, G. (2015). Deep learning. *nature, 521*(7553), 436-444.

Liu, Y., Ott, M., Goyal, N., Du, J., Joshi, M., Chen, D., . . . Stoyanov, V. (2019). Roberta: A robustly optimized bert pretraining approach. *arXiv preprint arXiv:1907*. Retrieved from https://arxiv.org/abs/1907.11692

Noy, N., Gao, Y., Jain, A., Narayanan, A., Patterson, A., & Taylor, J. (2019). Industry-scale Knowledge Graphs: Lessons and Challenges: Five diverse technology companies show how it's done. *Queue, 17*(2), 48-75.

Pieter, C., Davis, T., Gadepally, V., Kepner, J., Lipman, R., Lovitz, J., & Ouaknine, K. (2019). Redisgraph GraphBlas enabled graph database. *2019 IEEE International Parallel and Distributed Processing Symposium Workshops (IPDPSW)*, (pp. 285-286).

Radford, A., Narasimhan, K., Salimans, T., & Sutskever, I. (2018). *Improving language understanding by generative pre-training.* OpenAI. Retrieved from https://cdn.openai.com/research-covers/language-unsupervised/language_understanding_paper.pdf

Rae, J., Borgeaud, S., Cai, T., Millicanvvvv, K., Hoffmann, J., Song, F., . . . Powell, R. (2021). *Scaling language models: Methods, analysis & insights from training gopher.* DeepMind. arXiv preprint arXiv:2112.11446. Retrieved from https://arxiv.org/abs/2112.11446

Reimers, N., & Gurevych, I. (2019). Sentence-BERT: Sentence Embeddings using Siamese BERT-Networks. *Proceedings of the 2019 Conference on Empirical Methods in Natural Language Processing and the 9th International Joint Conference on Natural Language Processing (EMNLP-IJCNLP)* (pp. 3982-3992). Association for Computational Linguistics. doi:10.18653/v1/D19-1410

Robertson, S., & Zaragoza, H. (2009). The probabilistic relevance framework: BM25 and beyond. *Foundations and Trends® in Information Retrieval, 3*(4), 333-389.

Robinson, I., Webber, J., & Eifrem, E. (2015). *Graph databases: new opportunities for connected data.* O'Reilly Media, Inc.

Sanh, V., Debut, L., Chaumond, J., & Wolf, T. (2019). DistilBERT, a distilled version of BERT: smaller, faster, cheaper and lighter. *arXiv preprint arXiv:1910.01108.* Retrieved from https://arxiv.org/pdf/1910.01108.pdf

Scao, T., Fan, A., Akiki, C., Pavlick, E., Ilić, S., Hesslow, D., & Castagné, R. (2022). *Bloom: A 176b-parameter open-access multilingual language model.* BigScience Workshop. arXiv preprint arXiv:2211.05100.

Teng, S.-H., & Ho, S.-Y. (1996). Failure mode and effects analysis: an integrated approach for product design and process control. *International journal of quality & reliability management, 13*(5), 8-26.

Van Harmelen, F., Lifschitz, V., Porter, B., & eds. (2008). *Handbook of knowledge representation.* Elsevier.

Vaswani, A., Shazeer, N., Parmar, N., Uszkoreit, J., Jones, L., Gomez, A., . . . Polosukhin, I. (2017). Attention is all you need. *Advances in neural information processing systems*, *30.*

Wang, W., Wei, F., Dong, L., Bao, H., Yang, N., & Zhou, M. (2020). Minilm: Deep self-attention distillation for task-agnostic compression of pre-trained transformers. *Advances in Neural Information Processing Systems*, *33*, pp. 5776-5788.

Williams, A., Nangia, N., & Bowman, S. (2018). A Broad-Coverage Challenge Corpus for Sentence Understanding through Inference. *Proceedings of the 2018 Conference of the North American Chapter of the Association for Computational Linguistics: Human Language Technologies, Volume 1 (Long Papers)* (pp. 1112-1122). Association for Computational Linguistics.

Wu, Z., Liu, W., & Nie, W. (2021). Literature review and prospect of the development and application of FMEA in manufacturing industry. *The International Journal of Advanced Manufacturing Technology, 112*, 1409-1436.

Yang, Z., Dai, Z., Yang, Y., Carbonell, J., Salakhutdinov, R., & Le, Q. (2019). Xlnet: Generalized autoregressive pretraining for language understanding. *Advances in neural information processing systems*, *32.*

Zhu, Q., Gu, Y., Luo, L., Li, B., Li, C., Peng, W., . . . Zhu, X. (2021). When does Further Pre-training MLM Help? An Empirical Study on Task-Oriented Dialog Pre-training. *Proceedings of the*

*Second Workshop on Insights from Negative Results in NLP* (pp. 54-61). Association for Computational Linguistics.

# Appendix

## 1. Application Architecture

The goal of Fountain has been to be easily extensible to multiple manufacturing domains and applications. Hence it has been designed as a set of microservices deployable in a Kubernetes cluster as shown in fig 13. This provides the ability to test different knowledge representations and language models in parallel and quickly and reliably deploy different combinations in our cloud infrastructure. This provides a good option for experimentation as well as easy scaling when the load on the productive application increases.

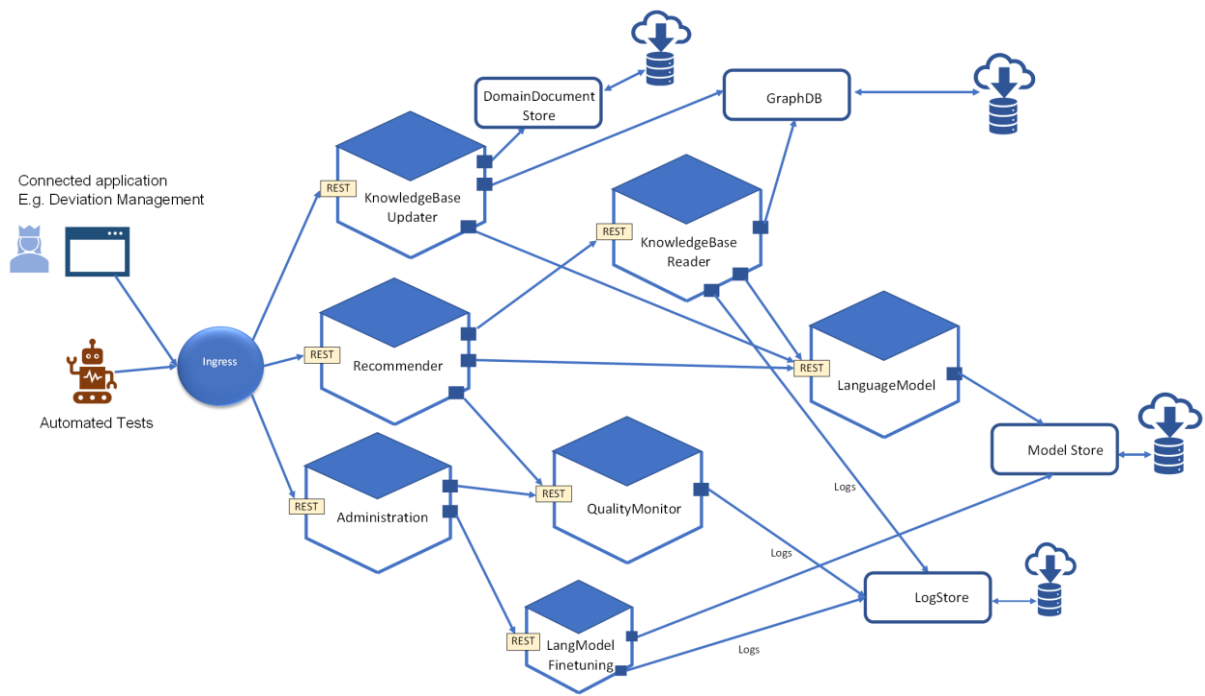

**Fig 13** Application architectures composed of microservices

## 2. Pretrained models evaluated

The following models available on the Hugging Face model hub have been used to evaluate the suitability of already available pretrained models for the domain specific semantic similarity task as shown in fig 7:

1. all-distilroberta-v1
    - Base model: distilroberta-base – distilled version of the RoBERTa-base model
    - Model card: https://huggingface.co/sentence-transformers/all-distilroberta-v1
2. all-roberta-large-v1
    - Base model: roberta-large
    - Model card: https://huggingface.co/sentence-transformers/all-roberta-large-v1
3. all-mpnet-base-v2
    - Base model: mpnet
    - Model card: https://huggingface.co/sentence-transformers/all-mpnet-base-v2

As per the model cards, the models have been trained on 1,124,818,467 sentence pairs.